\documentclass{article}
\usepackage{ijcai25}

\usepackage{times}
\usepackage{soul}
\usepackage{url}
\usepackage[hidelinks]{hyperref}
\usepackage[utf8]{inputenc}
\usepackage[small]{caption}
\usepackage{graphicx}
\usepackage{amsmath}
\usepackage{amsthm}
\usepackage{booktabs}
\usepackage{algorithm}
\usepackage{algorithmic}
\usepackage{natbib}
\usepackage{subcaption}
\usepackage[switch]{lineno}

\title{Federated Neural Architecture Search with Model-Agnostic Meta Learning}

\author{
Xinyuan Huang$^1$
\and
Jiechao Gao$^2$\thanks{Corresponding Author.}
\affiliations
$^1$University of Toronto\\
$^2$University of Virginia\\
\emails
victorxy.huang@mail.utoronto.ca,
jg5ycn@virginia.edu
}

\begin{document}

\maketitle

\begin{abstract}
    Federated Learning (FL) often struggles with data heterogeneity due to the naturally uneven distribution of user data across devices. Federated Neural Architecture Search (NAS) enables collaborative search for optimal model architectures tailored to heterogeneous data to achieve higher accuracy. However, this process is time-consuming due to extensive search space and retraining. To overcome this, we introduce FedMetaNAS, a framework that integrates meta-learning with NAS within the FL context to expedite the architecture search by pruning the search space and eliminating the retraining stage. Our approach first utilizes the Gumbel-Softmax reparameterization to facilitate relaxation of the mixed operations in the search space. We then refine the local search process by incorporating Model-Agnostic Meta-Learning, where a task-specific learner adapts both weights and architecture parameters (alphas) for individual tasks, while a meta learner adjusts the overall model weights and alphas based on the gradient information from task learners. Following the meta-update, we propose soft pruning using the same trick on search space to gradually sparsify the architecture, ensuring that the performance of the chosen architecture remains robust after pruning which allows for immediate use of the model without retraining. Experimental evaluations demonstrate that FedMetaNAS significantly accelerates the search process by more than 50\% with higher accuracy compared to FedNAS.
\end{abstract}

\section{Introduction}

With the proliferation of intelligent mobile devices, massive data is being generated at the user's end. In sectors like healthcare, it is important to leverage data across user devices for diagnostics while preserving their privacy. Therefore, federated learning (FL) \cite{FL_review} has been introduced, which enables devices across various platforms to collaboratively train a machine learning model without sharing local data, under the supervision of a central server. One of the fundamental challenges in FL is the non-identical and independent distribution (non-IID) of data across devices. In a non-IID setting, each device may have data that is not representative of the overall data distribution. This disparity can lead to models being biased towards the data characteristics of certain devices, resulting in reduced accuracy when aggregated \cite{fl_survey}. To mitigate this, various strategies such as data augmentation, local tuning, and structural adaptation have been proposed. Nonetheless, they rely on a predefined model architecture, which may not be optimal due to the opaque nature of the inherent data distribution. Consequently, researchers are compelled to explore multiple structures and fine-tune hyperparameters extensively. This search process can be costly in FL environment. \looseness=-1

Due to these challenges, Neural Architecture Search (NAS) \cite{elsken2019neuralarchitecturesearchsurvey} is introduced as a dynamic solution to target the non-IID data distribution. NAS is an automated method for optimizing the structure of a neural network. It constructs and evaluates a variety of model architectures based on predefined criteria. However, NAS is time-consuming due to the vastness of architecture search space, compounded with the two stage nature of its algorithm that involves searching and retraining. In FL, where achieving high accuracy must be balanced with stringent time constraints, this dual stage approach exacerbates existing solutions. Therefore, it is important to devise an NAS algorithm to effectively explore search spaces and accelerate the convergence of architectures.

To improve the accuracy in handling non-IID data and accelerate existing NAS based FL systems, we propose integrating Model-Agnostic Meta-Learning (MAML) \citep{maml} into a federated NAS framework. MAML is effective in rapidly adapting to new tasks with minimal data which aligns well with the dynamic and distributed nature of FL. This approach aims to reduce the iterative search time in traditional NAS, making it better suited to FL environments with non-IID data distributions. In our methodology, we begin by relaxing mixed operations within the search space using the Gumbel-Softmax reparameterization \cite{gumbel_trick} derived from the concrete distribution \cite{crv}, an approach that differs from the conventional softmax usage in DARTS \cite{darts}. This methods enables the targeted sparsification of the operations' weights. Subsequently, we adopt a dual-learner system consistent with MAML's framework for the local search process, enhancing the efficiency of model in adapting to new tasks. Given the interdependent nature of architecture parameters (alphas) and weights, both elements are jointly optimized by a base and a meta learner. During the base learning phase, alphas and weights are individually tailored for each task — represented by batches of data — while the meta learner aggregates the gradients' information from each task to update the overall model parameters. Following this meta update, we introduce a pruning technique that extends \cite{metanas}, leveraging the the Gumbel-Softmax reparameterization to sparsify the searched model during specific meta epochs. Specifically, we treat the input variables as combinations to facilitate the sparsification of individual input features. Through this implementation of soft pruning and the relaxation technique in the search space, the model architecture is able to converge during the search phase, eliminating the need for subsequent retraining. 
Currently, no existing studies have explored this combination of meta-learning, NAS, and FL. Therefore, our research aims to determine whether such integration can accelerate the processes of NAS without sacrificing model performance across non-IID data distributions. We propose a model by integrating MAML with NAS in a federated framework, FedMetaNAS. Our contributions are:  
\begin{itemize}
    \item Integration of MAML and NAS in FL: We have integrated MAML with NAS into local searching of model structures, allowing for meta-learning of both model weights and architectures simultaneously. 
    \item Advancement in Pruning and Sparsification: We have adapted and extended the pruning technique in \cite{metanas} with the Gumbel-Softmax reparameterization under FL setting, which eliminates retraining. 
    \item Experiments show that FedMetaNAS outperforms baseline methods in terms of accuracy. Additionally, it reduces the duration of the NAS process by over 50\% through accelerating model searching and eliminating the retraining stage required in FedNAS, demonstating the efficiency and effectiveness of our approach. \looseness=-1 
\end{itemize}

\section{Background}

In response to the challenge posed by non-IID in FL, NAS \cite{elsken2019neuralarchitecturesearchsurvey} emerges as a promising solution. It allows for the exploration and evaluation of various model structures, each tailored to unique data characteristics without manual intervention. This process involves three key components: search spaces, which define all possible architectures and operations; search strategies, such as evolutionary algorithms or gradient-based optimization, which are techniques used to discover high-performing architectures; and performance evaluation strategies. 

Among these search strategies, gradient-based NAS methods, such as DARTS \cite{darts}, are particularly advantageous in federated settings for its computational efficiency. Unlike evolutionary algorithms \cite{evol_alg, evol_alg2}, which require hundreds of GPU hours, gradient-based approaches optimize architectures in significantly fewer GPU hours \cite{elsken2019neuralarchitecturesearchsurvey}. DARTS employs a relaxation technique that transforms the selection of operations into a continuous distribution using a softmax function to allow for the soft selection of operations and integrating architectural decisions directly into the gradient descent. However, this methods introduces inconsistencies between the DARTS' loss function and its objective to maximize the expected performance of architectures sampled from this mixed operation distribution. Consequently, this necessitates post-architecture fine-tuning and retraining to reconcile these discrepancies.

Despite the computational advantages of gradient-based methods, their extensive time requirements still pose a significant limitation within FL, where time is crucial. To address this, MAML \cite{maml} offers a solution. In meta-learning, MAML enables models to learn and adapt quickly with only a few examples, continuing to refine as more data becomes available. It operates by optimizing a set of initial parameters that are effective across multiple tasks. This involves adjusting the model parameters specifically for a support task and evaluating the performance on a related query task. Similar to cross-validation, this allows MAML to enhance generalization across different tasks. By integrating MAML with NAS, the architecture optimization can be more effectively directed towards learning task-specific architectures, thereby reducing the overall time required for model searching and refinement.

\section{Related Works}
The integration of MAML with NAS presents a promising approach to overcome the limitations of FL in handling non-IID data. While the pairwise combinations of these three areas have been studied, the holistic integration of all three remains unexplored.

\subsection{NAS within FL}
Recent studies \cite{fednas, fedpnas} have demonstrated the capability of NAS to effectively optimize network architectures within federated settings, specifically addressing the challenges posed by non-IID data distributions that often lead to model divergence. For instance, FedNAS \cite{fednas} incorporates the MiLeNAS algorithm \cite{milenas} into FL to search for the optimal model architectures across distributed datasets. FedNAS employs a mixed-level optimization approach where the architecture and model weights are optimized simultaneously: architecture optimization utilizes both training and validation loss, while weight optimization uses only the training loss. Another application is seen in the work of FedPNAS \cite{fedpnas}, which introduces a progressive approach to neural architecture search within federated learning. FedPNAS progressively refines and evaluates candidate architectures across clients. However, both approaches come with drawbacks, such as extensive search and retraining times required to fully optimize the chosen model, which can be challenging in federated environments.

\subsection{MAML and NAS}
Another promising area is the integration of MAML with NAS. This combination seeks to utilize the quick adaptability of MAML with the structural optimization capabilities of NAS to create models that are capable of rapid adaptation to new environments. An example is T-NAS \cite{tnas}, which optimizes weights and architecture for different tasks separately, and then that aggregates these results to establish a initial point for further adaptation. Another model, MetaNAS \cite{metanas}, combines DARTS \cite{darts} with REPTILE \cite{reptile}, a variant of MAML, to learn new tasks from just a few examples. In a different approach, MTL-NAS \cite{mtlnas} integrates the principles of general-purpose multi-task learning (GP-MTL) \cite{mtl} with NAS. By segmenting the GP-MTL network into fixed task-specific backbones and employing a feature fusion scheme, MTL-NAS defines a task-agnostic search space that is compatible with NAS requirements. Yet, all of these methods are developed based on a centralized setting lacking the non-IID challenge of federated environments. 

\subsection{MAML within FL}
Another area of study is the application of MAML within FL. This combination focuses on MAML's ability to rapidly adapt models to diverse local datasets, addressing challenges such as personalization or data heterogeneity. For example, per-FedAvg \cite{perfedavg} incorporates MAML to derive an optimal global model that allows clients to achieve high local performance with minimal computational cost. 
On the other hand, FedMeta \cite{fedmeta} introduces an approach by distributing a shared meta-learner among clients. The meta-learner provides initial model parameters and learning rates enabling clients to efficiently adapt to their unique data distributions with fewer communication cost. However, both methods primarily focus on utilizing a fixed model architecture, which can be a significant limitation as they do not address the necessity of identifying and optimizing the best model architecture to suit a diverse and evolving data environment.

\section{Proposed Method}

\subsection{System Overview}

\begin{figure}[t]
\centering
\includegraphics[width=1\columnwidth]{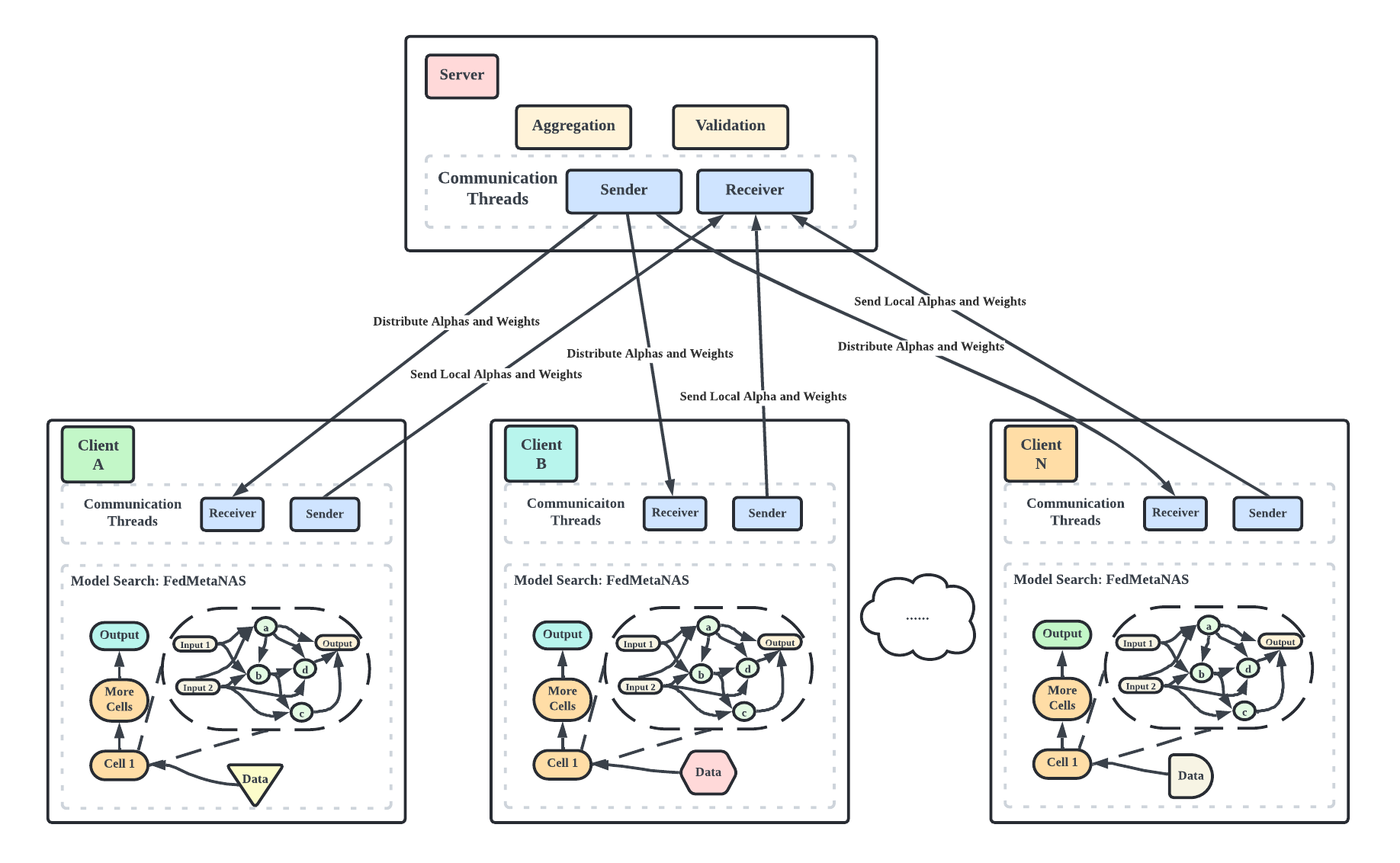} 
\caption{FedMetaNAS system overview}
\label{fig:system}\vspace{-0.1in}
\end{figure}


Referencing FedML \cite{fedml} and Plato \cite{plato}, we have designed our system framework as depicted in Figure \ref{fig:system}. Our architecture separates aggregation in a server, while the clients handle local searching tasks. Communication between the server and clients is facilitated through two threads: a sender and a receiver. They are responsible for transmitting the updated architecture parameters (alphas) and weights. When a new client joins, it receives the current alphas and weights from the server to start its local model development. The client then executes FedMetaNAS to search for an optimal model. Upon search completion, the client sends the updated parameters back to the server. The server aggregates these updates from chosen clients using FedAvg \cite{fedavg} and redistributes them for further iterations.


\subsection{Search Space}

\begin{figure}[t]
\centering
\includegraphics[width=0.85\columnwidth]{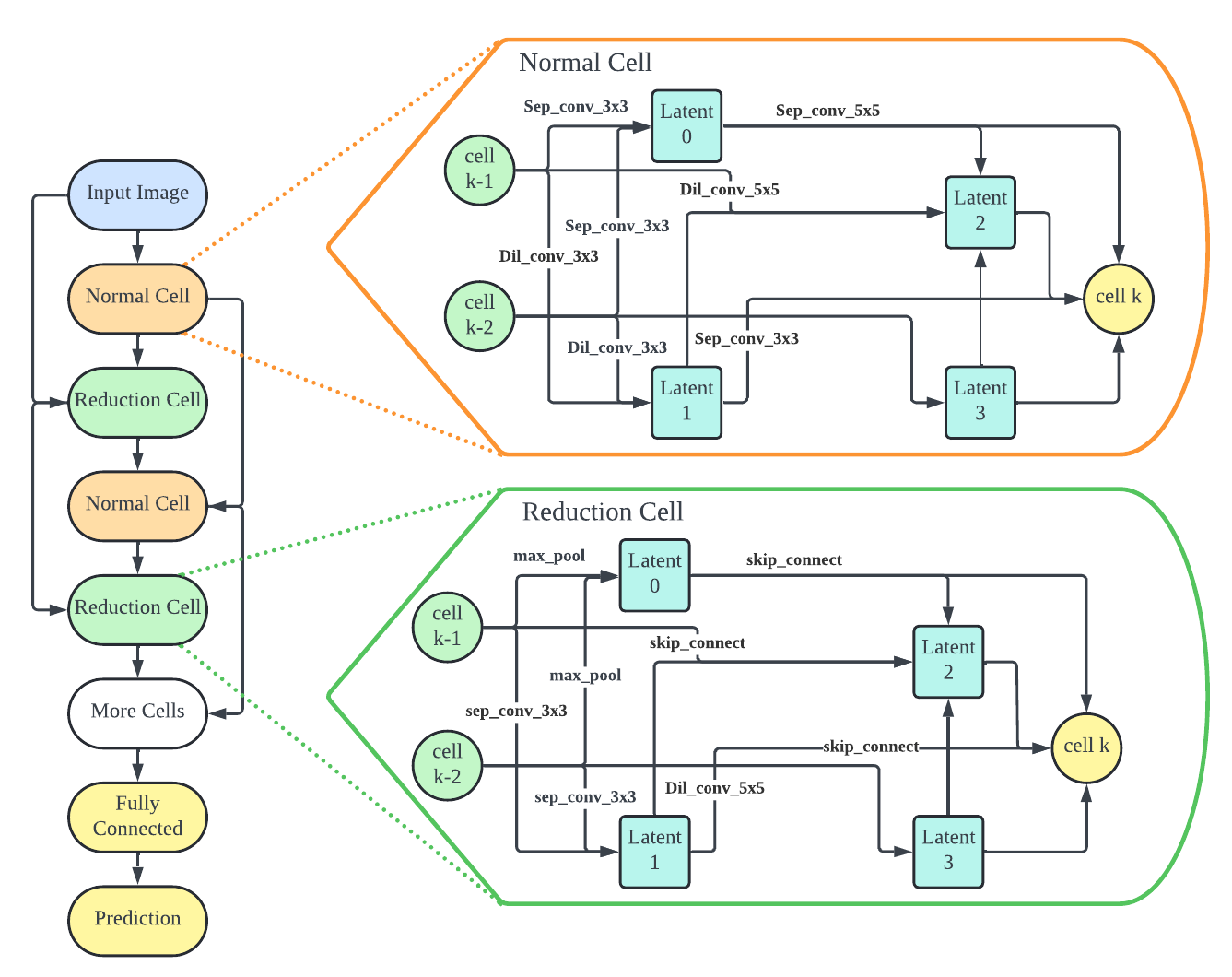} 
\caption{Sample search space}
\label{fig1:search_space}
\end{figure}

We designed our architecture search space based on DARTS \cite{darts}, as illustrated in Figure \ref{fig1:search_space}. Our approach utilizes a cell-based CNN architecture comprising two types of cells: normal cells, which preserve spatial dimensions, and reduction cells, which reduce dimensions through pooling. Each cell is modeled as a DAG with \(N\) nodes, where each node represents hidden feature representations interconnected via operations such as convolution or pooling. We have adopted the same set of primitive operations as DARTS.


Given the inconsistency between the loss function and the objective of DARTS, we integrate the Gumbel-Softmax reparameterization \cite{gumbel_trick} derived from the concrete distribution \cite{crv} for relaxing the operations which is inspired by SNAS \cite{snas}. The distribution is defined as:
\begin{equation}
\label{eqn:concreteRV}
    Z_k^{i,j} = \frac{\exp\left(\frac{\log \alpha_k^{i,j} + G_k^{i,j}}{\lambda}\right)}{\sum_{i=1}^n \exp\left(\frac{\log \alpha_i^{i,j} + G_i^{i,j}}{\lambda}\right)}
\end{equation}
where \(Z_k^{i,j}\) is the softened selection variable for operations at edge \(i,j\), \(\alpha_k^{i,j}\) are the architecture parameters, \(G_k^{i,j}\) is the \(k\)-th Gumbel random variable, and \(\lambda\) is the annealing temperature. Consequently, the overall mechanism is given by:
\begin{equation}
\label{eqn:darts_cell}
    x^{(j)} = \sum_{i<j} \sum_{o \in \mathcal{O}} Z_k^{i,j} \cdot o(x^{(i)}, w_{o}^{i,j})
\end{equation}
where \(x^{(j)}\) denotes the latent features at the \(j\)-th node, and \(o(x^{(i)}, w_{o}^{i,j})\) represents the operations applied between nodes \(i\) and \(j\). This modification enhances the diversification and stability of operation selection by aligning the objective and loss.

\subsection{Problem Definition}
In FL, we consider a scenario with \( K \) clients, each possessing a non-IID dataset \( D_k \). The collaborative training objective for a neural network is defined as:
\begin{equation}
    \min_{w} F(w) = \min_{w} \sum_{k=1}^K \frac{n_k}{n} \cdot f_k(w)
\end{equation}
where \( w \) represents the model weights, and \( f_k(w) \) denotes the local objective function for client \( k \) with fixed architectural parameters.

In contrast, FedNAS aims to concurrently optimize both the network architecture, \( \alpha \), and the weights, \( w \), as expressed in:
\begin{equation}
\label{eqn:fednas}
    \min_{w, \alpha} F(w, \alpha) = \min_{w, \alpha} \sum_{k=1}^K \frac{n_k}{n} \cdot f_k(w, \alpha)
\end{equation}

Building upon this, our approach seeks to optimize both the architecture \( \alpha \) and the weights \( w \) by integrating MAML from a task-oriented perspective. Adhering to the same objective in equation \ref{eqn:fednas}, we formalize our local objective as:
\begin{equation}
    \min_{w, \alpha} \frac{1}{n} \sum_{k=1}^K f_k(w-\eta_w \nabla_w f_k(w, \alpha), \alpha - \eta_a \nabla_\alpha f_k(w, \alpha))
\end{equation}
where \( \eta_w \) and \( \eta_\alpha \) are distinct learning rates for the weights and architecture respectively. \( \nabla_w f_i(w, \alpha) \) and \( \nabla_\alpha f_i(w, \alpha) \) represent the respective partial derivatives. 

Within MAML, the model utilizes both a base learner and a meta learner to facilitate updates at the task and meta levels respectively. Similarly, our model employs task-specific optimizer for the inner updates and a meta optimizer for external meta adjustments.

The task optimizer refines both the architecture \( \tilde{\alpha} \) and the weights \( \tilde{w} \) based on support tasks \( T_s^k \) from local dataset \( D^k \), as shown below:
\begin{equation}
\label{eqn:task_optim}
\left\{
\begin{array}{l}
\hat{w}_{m+1}^k = \hat{w}_m^k - \eta_{\hat{w}}^{\text{task}} \nabla_{\hat{w}_m^k} L(T_s^k; \hat{w}_m^k, \hat{\alpha}_m^k) \\
\hat{\alpha}_{m+1}^k = \hat{\alpha}_m^k - \eta_{\hat{\alpha}}^{\text{task}} \nabla_{\hat{\alpha}_m^k} L(T_s^k; \hat{w}_{m+1}^k, \hat{\alpha}_m^k)
\end{array}
\right.
\end{equation}
where \( m \) denotes the number of inner update steps within the loop. After \( M \) steps, the meta optimizer performs updates to establish a good starting point for subsequent adaptations:
\begin{equation}
\label{eqn:meta_optim}
\left\{
\begin{array}{l}
w^k = w^k - \eta_{w}^{\text{meta}} \nabla_{w^k} L(T_q^k; w_M^k, \alpha_M^k) \\
\alpha^k = \alpha^k - \eta_{\alpha}^{\text{meta}} \nabla_{\alpha^k} L(T_q^k; w_M^k, \alpha_M^k)
\end{array}
\right.
\end{equation}
Here, \( T_q^k \) refers to the query tasks drawn from the local datasets. We propose that this formulation can enhance the efficacy of FL. The complete algorithm of FedMetaNAS is detailed in Algorithm \ref{alg:FedMetaNAS}.

\begin{algorithm}
\caption{FedMetaNAS Algorithm}
\label{alg:FedMetaNAS}
\begin{algorithmic}[1]
\STATE \textbf{Initialization:} Initialize \( w_0 \) and \( \alpha_0 \); Select \( K \) clients for update; R communication round;
\STATE \textbf{Server does:}
\FOR{\( t = 0 \) to \( R-1 \)}
    \FORALL{\( k \)}
        \STATE \( (w_{t+1}^k, \alpha_{t+1}^k) \leftarrow \text{Client Search}(k, w_t, \alpha_t) \)
    \ENDFOR
    \STATE \( w_{t+1} \leftarrow \sum_{k=1}^K \frac{N_k}{N} w_{t+1}^k \)
    \STATE \( \alpha_{t+1} \leftarrow \sum_{k=1}^K \frac{N_k}{N} \alpha_{t+1}^k \)
\ENDFOR

\STATE \textbf{Client Search}(\( w, \alpha \))\textbf{:}
\FOR{\( e = 1 \) to \( E \)}
    \STATE Sample support and query task \( T_s^k \) and \( T_q^k \) from \( D_k \)
    \FOR{ \( m = 1\) to \( M \)}
        \STATE Compute \( L(T_s^k; \tilde{w}_m^k, \tilde{\alpha}_m^k) \)
        \STATE Update \( \tilde{w}_m^k \) and \(\tilde{\alpha}_m^k\) using Equation \ref{eqn:task_optim}
    \ENDFOR
    \STATE Compute \( L(T_q^k; w^k, \alpha^k) \)
    \STATE Update \( w^k\) and \( \alpha^k\) using Equation \ref{eqn:meta_optim}
    \STATE Model Pruning
\ENDFOR
\STATE \textbf{return} \( w, \alpha \) to server

\end{algorithmic}
\end{algorithm}

\subsection{Input Nodes Pruning}
\label{sc:pruning}

While sparsifying the operations enhances stability during architectural search, previous study \cite{metanas} indicates that this approach alone is insufficient, as it generates only a one-hot mixture of parameters within each mixed operation (the inner sum in Equation \ref{eqn:darts_cell}). The aggregation of mixed operations within a cell remains compact (the outer sum in Equation \ref{eqn:darts_cell}). To address this, we improve upon the method for soft pruning of the input nodes as proposed in \cite{metanas}. This method extends the concept of sparsification to the architecture's structure by employing a softmax function to relax the mixed operations. Specifically, the solution permits \(k\) combinations of mixed operations to facilitate the selection of multiple inputs as DARTS requires 2.

However, instead of relying on the softmax function, we employ the same Gumbel-Softmax reparameteriation \cite{gumbel_trick} described in Equation \ref{eqn:concreteRV} to sparsify the input nodes:
\begin{equation}
    x^j = \sum_{i \in I} Z_k^{i,j} \cdot (MixedOp(x^{i_1}) + \ldots + MixedOp(x^{i_k})),
\end{equation}
where \(I\) represents all possible combinations of \(k\) input nodes, and \(Z_k^{i,j}\) denotes the concrete random variable from Equation \ref{eqn:concreteRV}. At the end of certain meta update, we simply keep all operations and input nodes with corresponding weight \(\alpha \) larger than some threshold, while all others are pruned. This ensures that the architecture selected during the search process can be pruned without a performance degradation.

\subsection{Optimization}
There are several computational challenges in the problem formulation as it generates a four-level optimization: bilevel for each optimizer. 
Specifically, the inclusion of a second-order derivative in Equation \ref{eqn:meta_optim} complicates the optimization due to its high computational cost. To address this, we adopt a first-order approximation of the second-order derivative, as discussed in \cite{darts}, which significantly reduces the overhead. 

Moreover, the simultaneous optimization of alphas and weights intensifies the computational demands. It often leads to extensive memory consumption, especially when the task update steps, \(M\), is large. To solve this, we have implemented a joint optimization for both weights and alphas, similar to the approach used by \cite{tnas}, where both parameters are treated uniformly during backpropagation. This simplifies the optimization landscape and reduces the frequency of backpropagation.

The revised optimization steps are formalized as follows:
For the task updates, we update the parameters using the gradient descent rule:
\begin{equation}
 \begin{aligned}
    (\hat{w}_{m+1}^k, \hat{\alpha}_{m+1}^k) = (\hat{w}_{m}^k, \hat{\alpha}_{m}^k) - \boldsymbol{\eta^{\text{task}}} \\ \cdot \boldsymbol{\nabla}_{(\hat{w}_{m+1}^k, \hat{\alpha}_{m+1}^k)} L(T_s^k; \hat{w}_{m+1}^k, \hat{\alpha}_m^k)
\end{aligned}
\end{equation}
 where \( \boldsymbol{\eta^{\text{task}}} \) represents the matrix of learning rates for both \(\hat{w}_{m+1}^k\) and \(\hat{\alpha}_{m+1}^k\). Similarly for the meta update we apply:
 \begin{equation}
 \begin{aligned}
    ({w}_{m+1}^k, {\alpha}_{m+1}^k) = ({w}_{m}^k, {\alpha}_{m}^k) - \boldsymbol{\eta^{\text{meta}}} \\ \cdot \boldsymbol{{\nabla}}_{({w}_{m+1}^k, {\alpha}_{m+1}^k)} L(T_s^k; {w}_{m+1}^k, {\alpha}_m^k)
\end{aligned}
\end{equation}
where the gradient, \({\nabla}_{({w}_{m+1}^k, {\alpha}_{m+1}^k)} L(T_s^k; {w}_{m+1}^k, {\alpha}_m^k)\), is approximated using a first-order method.

\section{Experiments}

This section presents a comparative evaluation of our proposed algorithm against current state of the art federated learning methods and prior federated NAS approaches.

\subsection{Experimental Settings}

To evaluate the effectiveness of our methods, we employ CNN search spaces (DARTS \cite{darts}) and compare our method against fixed MobileNetV3-Large \cite{mobilenet} model since our proposed algorithm targets gradient-based search space in particular. MobileNetV3 belongs to a series of CNNs designed through NAS specifically for mobile phone CPUs that aligns well with FL's computational constraints.

We select the following image classification datasets for our training tasks: CIFAR10, CIFAR100, and MNIST, each consisting of 60,000 images. For each dataset, 50,000 images were distributed among the clients, and 10,000 were reserved for centralized server testing. Each client was allocated 600 images for local training and another 600 for evaluation.

Our experiments are operated under non-IID conditions to mimic real-world federated learning environments. Non-IID data distribution was modeled either as label-skewed or using a Dirichlet distribution with parameter $Dir_{\alpha}$. A small $Dir_{\alpha}$ indicates a high degree of data heterogeneity \cite{fl_survey}.

Our evaluation metrics include the overall model accuracy on the centralized server and the average testing accuracy across clients. Additionally, we measured the total computational time required to reach the reported accuracies. Our experimental design incorporated a global model search framework where multiple clients collaboratively optimized a unified model. All methods are trained until they fully converged using over 1000 communication rounds. We employ a cross-device configuration, selecting 5 out of 100 clients in each round to perform 5 epochs of local training. The experimental conditions, including method fine-tuning and random seed settings, are controlled with results reported over five independent runs. All computational experiments are conducted using NVIDIA A100 GPUs.

\subsection{Against Conventional Federated Methods}
\begin{figure}[t]
  \centering
  \begin{minipage}{0.48\columnwidth}
    \centering
    \includegraphics[width=\linewidth]{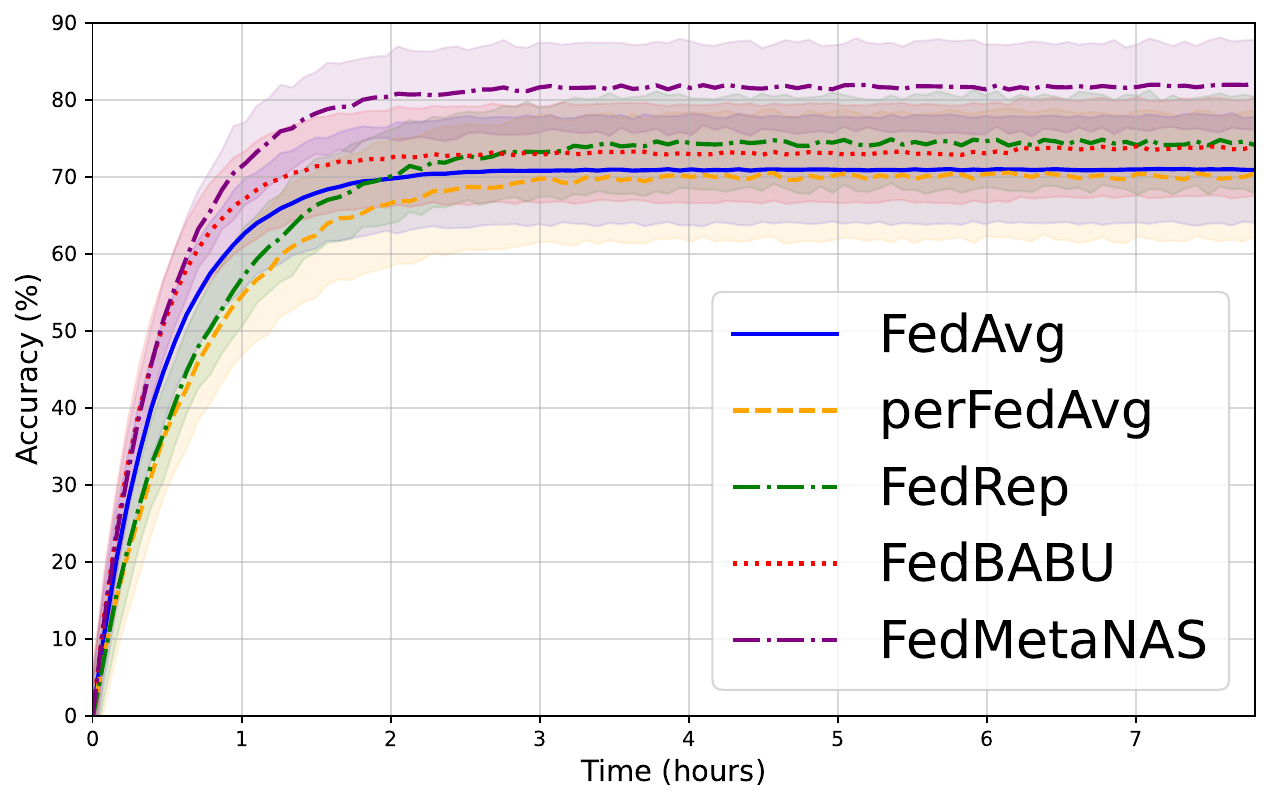}
    \captionsetup{justification=centering}
    \subcaption{CIFAR10 ($Dir_{\alpha} = 0.1$)}
    \label{fig:cifar10_conventional}
  \end{minipage}\hfill
  \begin{minipage}{0.48\columnwidth}
    \centering
    \includegraphics[width=\linewidth]{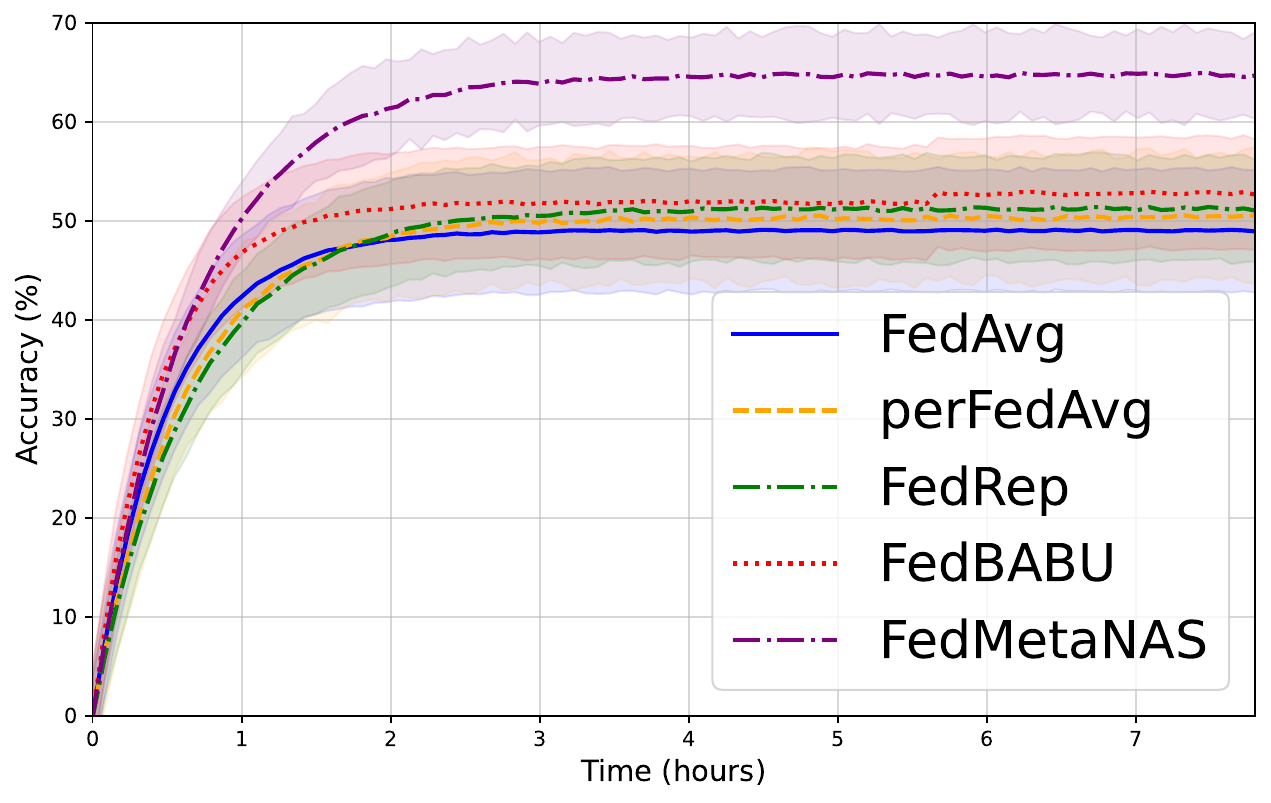}
    \captionsetup{justification=centering}
    \subcaption{CIFAR100 ($Dir_{\alpha} = 0.1$)}
    \label{fig:cifar100_conventional}
  \end{minipage}
  \caption{Test Accuracy with Standard Deviation over time on CNN}
  \label{fig:conventional_result}
\end{figure}

\begin{table}[t]
\centering
\scalebox{0.9}{
\begin{tabular}{@{}lcccc@{}}
\toprule
\textbf{Datasets} & \multicolumn{2}{c}{Cifar10 $Dir_{\alpha}=0.1$} & \multicolumn{2}{c}{Cifar100 $Dir_{\alpha}=0.1$} \\ 
\cmidrule(lr){2-3} \cmidrule(lr){4-5}
\textbf{Methods} & \textbf{Accuracy (\%)} & \textbf{Time} & \textbf{Accuracy (\%)} & \textbf{Time} \\ \midrule
FedAvg        & $70.93 \pm 7.01$ & 6.81 & $49.02 \pm 6.21$& 6.17  \\
per-FedAvg        & $70.13 \pm 8.24$ & 7.6 & $50.31 \pm 6.48$ & 7.69  \\
FedRep    & $74.51 \pm 6.14$ & 7.47 & $51.23 \pm 5.28$ & 7.33  \\
FedBABU  & $73.67 \pm 6.35$ & 7.07 & $52.77 \pm 5.64$ & 6.85 \\
\cmidrule(){1-5}
FedMetaNAS     & $81.65 \pm 5.72$ & 7.12 & $64.75 \pm 4.47$ & 7.35  \\ 
\bottomrule
\end{tabular}}
\caption{Performance comparison between conventional federated methods on various datasets}
\label{tb:conventional}\vspace{-0.1in}
\end{table}

We first assess our proposed method against other FL techniques by measuring the general testing accuracy on the server to validate the effectiveness of our searched architecture. As a baseline, we use FedAvg \cite{fedavg} and per-FedAvg \cite{perfedavg}, an enhancement of FedAvg which inspired the incorporation of MAML into our process. Additionally, we compare our approach with FedRep \cite{fedrep}, which separates representation learning and personalization, and FedBABU \cite{fedbabu}, which employs Bayesian updates to refine global model parameters. These methods are chosen for their leading roles in addressing non-IID data challenges and their proven superior performance over earlier techniques such as FedProx \cite{fedprox}, Ditto \cite{ditto}, and SCAFFOLD \cite{scaffold}. These conventional methods are trained using a fixed CNN model, MobileNetV3-Large, across datasets. The results are summarized in Table \ref{tb:conventional} and Figure \ref{fig:conventional_result}.

The results in Table \ref{tb:conventional} demonstrate that FedMetaNAS significantly outperforms conventional federated learning methods, achieving higher accuracy across both datasets under non-IID conditions while maintaining comparable computational time. Although FedRep and FedBABU show improvements over FedAvg and per-FedAvg, they are constrained by the limitation of using fixed model architectures. Furthermore, although methods like FedAvg and FedBABU reach their peak accuracy faster, FedMetaNAS surpasses their performance significantly at comparable time points as shown in Figure \ref{fig:conventional_result}. The superior performance of FedMetaNAS can be attributed to the incorporation of NAS, which identifies optimal architectures tailored to heterogeneous client data that effectively address the limitations of fixed-model approaches. Additionally, the integration of the pruning mechanism eliminates the need for retraining which is a common contributor to the long  time of NAS methods. This allows FedMetaNAS to efficiently explore the search space while maintaining computational efficiency.

\subsection{Against Prior Federated NAS methods}
\label{sc:fedNAS}

\begin{figure}[t]
  \centering
  \begin{minipage}{0.48\columnwidth}
    \centering
    \includegraphics[width=\linewidth]{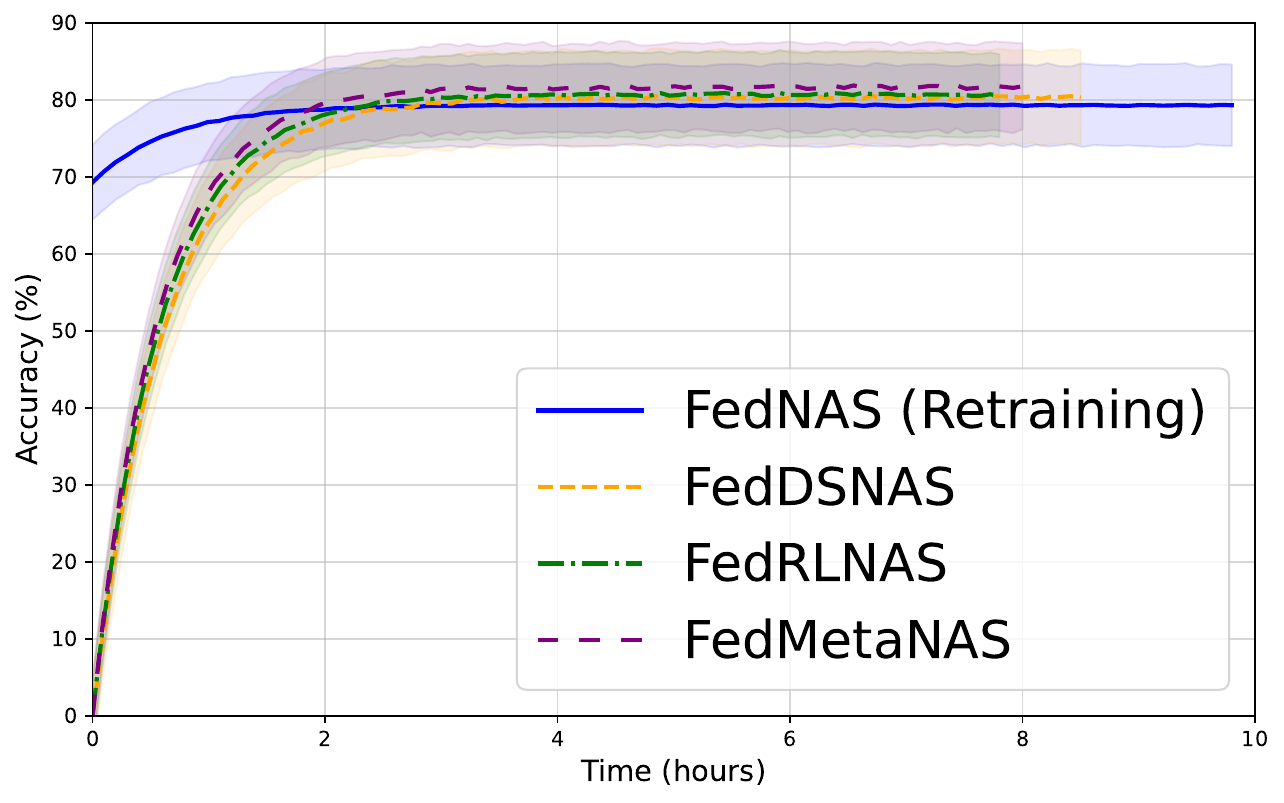}
    \captionsetup{justification=centering}
    \subcaption{CIFAR10 ($Dir_{\alpha} = 0.1$)}
    \label{fig:cifar10_conventional}
  \end{minipage}\hfill
  \begin{minipage}{0.48\columnwidth}
    \centering
    \includegraphics[width=\linewidth]{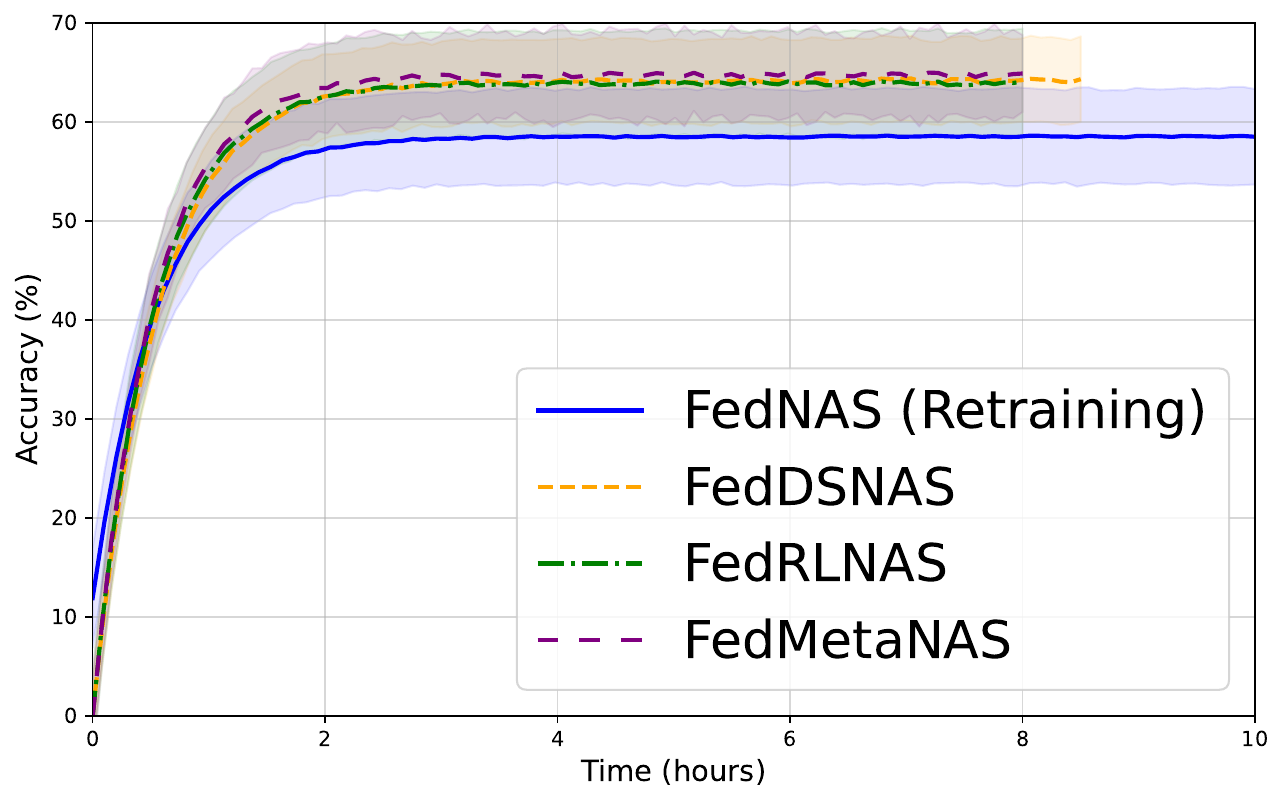}
    \captionsetup{justification=centering}
    \subcaption{CIFAR100 ($Dir_{\alpha} = 0.1$)}
    \label{fig:cifar100_conventional}
  \end{minipage}
  \caption{Test accuracy with standard deviation over time in DARTS search space}
  \label{fig:nas_result}
\end{figure}

\begin{table}[t]
\centering
\scalebox{0.9}{
\begin{tabular}{@{}lcccc@{}}
\toprule
\textbf{Datasets} & \multicolumn{2}{c}{Cifar10 $Dir_{\alpha}=0.1$} & \multicolumn{2}{c}{Cifar100 $Dir_{\alpha}=0.1$} \\ 
\cmidrule(lr){2-3} \cmidrule(lr){4-5}
\textbf{Methods} & \textbf{Accuracy (\%)} & \textbf{Time} & \textbf{Accuracy (\%)} & \textbf{Time} \\ \midrule
FedNAS         & $79.31 \pm 5.23$ & 19.47 & $58.52 \pm 4.86$ & 20.42  \\
FedDSNAS        & $80.27 \pm 6.16$ & 8.08 & $64.13 \pm 4.28$ & 8.19  \\
FedRLNAS    & $80.71 \pm 5.48$ & 7.23 & $63.89 \pm 5.33$ & 7.51  \\
\cmidrule(){1-5}
FedMetaNAS     & $81.65 \pm 5.72$ & 7.12 & $64.75 \pm 4.47$ & 7.35  \\ 
\bottomrule
\end{tabular}}
\caption{Performance comparison between federated NAS methods on various datasets}
\label{tb:nas}\vspace{-0.1in}
\end{table}

In this subsection, we benchmark FedMetaNAS against prior Federated NAS methods that utilize a general supernet architecture distributed across all clients focusing on global testing accuracy. Our comparison includes FedNAS \cite{fednas}, the foundational work in Federated NAS, which demonstrated the feasibility of applying NAS within a federated learning setting; FedDSNAS \cite{fedpnas}, a federated adaptation of DSNAS \cite{dsnas} designed to improve efficiency through gradient-based optimization; and FedRLNAS \cite{fedrlnas}, which leverages reinforcement learning to dynamically optimize architectures across distributed nodes. All methods employ the DARTS search space for consistency with the results presented in Table \ref{tb:nas} and Figure \ref{fig:nas_result}.

As shown in Table \ref{tb:nas}, FedMetaNAS outperforms prior Federated NAS methods in both accuracy and convergence efficiency. It achieves the highest accuracy of 81.65\% on CIFAR10 and 64.75\% on CIFAR100, surpassing all baselines while maintaining a slightly faster convergence rate. FedNAS, the  pioneer in Federated NAS methods, follows a multi-stage NAS process involving warm-up and fine-tuning phases, which significantly extend its training time. For understandability, only the retraining phase of FedNAS is shown in Figure \ref{fig:nas_result}. In contrast, FedDSNAS, FedRLNAS, and FedMetaNAS employ single-stage NAS approaches, which directly select architectures during the search phase, thereby reducing overall time. Among these, FedMetaNAS demonstrates superior performance by converging faster and achieving higher accuracy on both datasets which showcase its efficiency and effectiveness through the integration of MAML and pruning.

\subsection{Results of Soft Pruning}
\begin{table}[]
\centering
\scalebox{0.9}{
\begin{tabular}{@{}lcccccc@{}}
\toprule
Datasets & \multicolumn{2}{c}{Cifar10} & \multicolumn{2}{c}{MNIST} & \multicolumn{2}{c}{Cifar100} \\ 
\cmidrule(lr){2-3} \cmidrule(lr){4-5} \cmidrule(lr){6-7}
Pruning & Before & After & Before & After & Before & After \\ 
\midrule
FedNAS    & 79.31 & 67.26 & 98.60 & 83.46 & 58.52 & 11.26 \\
FedMetaNAS & 81.65 & 81.65 & 98.67 & 98.66 & 64.75 & 64.753 \\
\bottomrule
\end{tabular}}
\caption{Results of pruning on various datasets}
\label{tb:pruning-results}\vspace{-0.1in}
\end{table}

We assessed the impact of soft pruning in FedMetaNAS relative to FedNAS, which contains a variant of DARTS that still retains a hard pruning mechanism. Specifically, we conducted the search phase on CIFAR10, MNIST, and CIFAR100, and a significant decline in FedNAS's test accuracy is observed after search phase when transitioning from the searched architecture to the pruned model. The results which is shown in Table \ref{tb:pruning-results} shows a significant performance deterioration across all datasets, with the most severe drop noted in CIFAR100. We attribute this degradation to the premature convergence of model architecture during the search phase, particularly under the complex conditions of CIFAR100. In contrast, our algorithm maintains stable performance after pruning which benefits from the node pruning and annealing processes during the architecture search phase (section \ref{sc:pruning}). 

\subsection{Results on Non-IID Data Distribution}

\begin{table}[t]
\centering
\scalebox{0.8}{
\begin{tabular}{@{}lcccc@{}}
\toprule
Non-IID    & \(Dir_{\alpha}\) = 0.3 & \(Dir_{\alpha}\) = 0.1 & \(\tau\) = 8 & \(\tau\) = 4 \\ \midrule
FedNAS    & $80.32 \pm 5.64$ & $79.31 \pm 5.23$ & $78.22 \pm 5.86$ & $74.34 \pm 6.94$ \\
FedDSNAS  & $81.44 \pm 5.89$ & $80.27 \pm 6.16$ & $79.19 \pm 6.01$ & $77.86 \pm 6.58$ \\
FedRLNAS  & $81.25 \pm 5.54$ & $80.71 \pm 5.48$ & $80.48 \pm 5.76$ & $78.59 \pm 6.47$ \\
\cmidrule(){1-5}
FedMetaNAS & $82.43 \pm 5.29$ & $81.65 \pm 5.72$ & $81.08 \pm 5.68 $  & $78.73 \pm 6.23$\\ 
\bottomrule
\end{tabular}}
\caption{Performance comparison across non-IID trials in Cifar10}
\label{tb:iid_results}\vspace{-0.2in}
\end{table}

We further evaluated the efficacy of FedMetaNAS on CIFAR10 dataset under various non-IID conditions. Our baseline models include prior federated NAS methods mentioned in subsection \ref{sc:fedNAS}. We varied the parameter \(\alpha\) in the Dirichlet distribution, along with the number of labels in the clients' local dataset to explore different degrees of data heterogeneity. Specifically, we tested configurations with \(\alpha = 0.3\) and \(\alpha=0.1\) where each client has all classes but distributed unevenly and label skews where each client contains only 4 or 8 ($\tau = 4$ or $\tau = 8$) classes with each class contain almost the same amount of samples. 

Across all configurations, FedMetaNAS consistently outperforms the baseline methods, achieving the highest accuracy in each scenario. Under $\alpha = 0.3$, FedMetaNAS achieves $82.43\%$, a 0.99\% improvement over FedDSNAS, the next best method, and a 2.11\% improvement over FedNAS. Similarly, under more extreme heterogeneity with $\alpha = 0.1$, FedMetaNAS achieves $81.65\%$, surpassing FedDSNAS and FedRLNAS by 1.38\% and 0.94\% respectively. FedMetaNAS’s advantage is also evident in the label skew experiments, where it achieves $81.08\%$ at $\tau = 4$ and $78.73\%$ at $\tau = 4$, consistently outperforming other methods. Under $\tau = 4$ which is a highly challenging scenario, FedMetaNAS surpasses FedDSNAS by 0.87\% and FedRLNAS by 0.14\% demonstrating its robustness to extreme non-IID conditions. Notably, FedMetaNAS also exhibits relatively lower variance across trials indicating greater stability and reliability compared to its counterparts. These results illustrates the effectiveness of FedMetaNAS’s single-stage NAS approach combined with MAML in dynamically adapting architectures to diverse client data distributions.

\subsection{Personalization Performance}

\begin{table}[t]
\centering
\scalebox{0.75}{
\begin{tabular}{@{}lcccc@{}}
\toprule
\textbf{Datasets} & \multicolumn{2}{c}{Cifar10 $Dir_{\alpha}=0.1$} & \multicolumn{2}{c}{Cifar100 $Dir_{\alpha}=0.1$} \\ 
\cmidrule(lr){2-3} \cmidrule(lr){4-5}
\textbf{Methods} & \textbf{Accuracy (\%)} & \textbf{Time} & \textbf{Accuracy (\%)} & \textbf{Time} \\ \midrule
FedNAS         & $78.69 \pm 5.76$ & 19.47 & $58.11 \pm 4.86$ & 20.42  \\
FedMetaNAS (ours)    & $81.24 \pm 5.42$ & 7.12 & $64.45 \pm 4.47$ & 7.35  \\
\cmidrule(lr){1-5}
FedPNAS    & $81.01 \pm 5.79$ & 10.29 & $63.65 \pm 4.81$ & 10.02  \\
meta-FedPNAS (ours) & $81.87 \pm 5.57$ & 9.57 & $64.98 \pm 4.96$ & 9.66  \\ 
\cmidrule(lr){1-5}
PerFedRLNAS    & $81.72 \pm 5.32$ & 7.29 & $64.71 \pm 4.74$ & 7.18  \\
meta-PerFedRLNAS (ours) & $82.36 \pm 5.50$ & 7.15 & $65.16 \pm 4.17$ & 6.97  \\ 
\bottomrule
\end{tabular}}
\caption{Parallel performance comparison between federated NAS methods on various datasets}
\label{tb:personalized}\vspace{-0.1in}
\end{table}

In addition to global test accuracy, we also explored the effectiveness of our framework within personalized settings by measuring average local test accuracies. Our analysis involved three parallel comparisons to distinguish between different federated NAS strategies. The first comparison examines FedMetaNAS against FedNAS, where both methods train a single generalized model distributed to all clients. This is ideal for scenarios requiring uniform application, such as large healthcare systems. The second comparison assesses 
 FedPNAS and meta-FedPNAS. FedPNAS personalizes NAS by distributing tailored subnets, while meta-FedPNAS incorporates our proposed MAML and pruning techniques for further optimization. Finally, we compared perFedRLNAS \cite{perfedrlnas}, an enhanced version of FedRLNAS \cite{fedrlnas} that personalizes the architecture by training a supernet and distributing tailored subnets to each client using reinforcement learning, suitable for applications where client-specific optimization can significantly enhance performance, such as in mobile applications. Against this, we used meta-perFedRLNAS, which we incorporated the proposed MAML and pruning techniques into the optimization of the search process.

As Table \ref{tb:personalized} shows, FedMetaNAS demonstrates significant improvements over FedNAS, achieving $81.24\%$ on CIFAR10 and $64.45\%$ on CIFAR100, compared to $78.69\%$ and $58.11\%$ respectively, while also reducing training time. These results illustrate the ability of our proposed algorithm to adapt architectures more efficiently by incorporating MAML and node pruning under a personalized setting. When compared to FedPNAS and PerFedRLNAS, which are specifically designed for personalized FL, our proposed FedMetaNAS also performs competitively in terms of average client test accuracy. Slightly improved versions, meta-FedPNAS and meta-PerFedRLNAS, achieve even better performance, with meta-PerFedRLNAS delivering the highest average client accuracy at $82.36\%$ on CIFAR10 and $65.16\%$ on CIFAR100. This highlights the benefits of MAML and pruning in the architecture search process for client-specific optimization, as they provide an efficient optimization of the search space. Overall, these results confirm the versatility of FedMetaNAS and the effectiveness of MAML and node pruning under the DARTS search space in both generalized and personalized federated learning scenarios.

\subsection{Limitations}

Despite the results of FedMetaNAS, there are some limitations in our current study. First, we did not explicitly optimize for and evaluate computational overhead and communication efficiency, which are critical attributes in FL scenarios. These factors often play a significant role in determining the viability of FL methods particularly in resource constrained environments. Additionally, our experiments were limited to the DARTS search space and we did not explore the Vision Transformer (ViT) search spaces. This is due to ViT based search space \cite{nasvit} rely on evolutionary algorithms which are incompatible with our proposed methods in gradient based optimization. Furthermore, existing research on meta-learning and NAS for ViT is limited, and no prior work has investigated this combination in an FL setting. Given these gaps, our future work will focus on exploring the transferability and adaptability of our method to alternative search spaces, such as ViT, to enhance its applicability across a broader range of tasks.

\section{Conclusion}

In this paper, we introduce FedMetaNAS, a NAS-based FL system that incorporates MAML and soft pruning to design a global model with high accuracy and reduced time, specifically addressing non-IID data distributions. We began by employing a the Gumbel Soft trick to relax the search space, enabling sparsification of mixed operations in the search phase. Subsequently, we implemented a dual-learner system in the local search, optimizing both architecture parameters and weights simultaneously. This system comprises a task learner, tailored to individual tasks, and a meta learner that utilizes gradient information from the task learner to refine the global model. We also applied soft pruning to input nodes, using the same relaxation trick to encourage the model architecture to converge to a one-hot representation during the search process. Comparative experiments under non-IID conditions demonstrated that FedMetaNAS outperforms baseline models significantly reducing total process time by over 50\% relative to FedNAS.

\appendix




\clearpage
\bibliographystyle{named}
\bibliography{ijcai25}

\end{document}